\documentclass[conference]{IEEEtran}
\IEEEoverridecommandlockouts

\usepackage{algorithm}
\usepackage{algpseudocode}

\usepackage{cite}
\usepackage{pgfplots}
\pgfplotsset{compat=1.3} 
\usepackage{tikz}
\usepackage{subcaption}

\usepackage{dblfloatfix}

\usepackage{amsmath,amssymb,amsfonts}
\usepackage{stmaryrd}
\usepackage{graphicx}
\usepackage{textcomp}
\usepackage{xcolor}
\def\BibTeX{{\rm B\kern-.05em{\sc i\kern-.025em b}\kern-.08em
    T\kern-.1667em\lower.7ex\hbox{E}\kern-.125emX}}

\begin{document}

\title{Leveraging Structured Pruning of \\Convolutional Neural Networks
}
\author{
    \IEEEauthorblockN{Hugo Tessier\IEEEauthorrefmark{1}\IEEEauthorrefmark{2}, Vincent Gripon\IEEEauthorrefmark{2}, Mathieu Léonardon\IEEEauthorrefmark{2}, Matthieu Arzel\IEEEauthorrefmark{2}, David Bertrand\IEEEauthorrefmark{1}, Thomas Hannagan\IEEEauthorrefmark{1}}
    \IEEEauthorblockA{\IEEEauthorrefmark{1}\textit{Stellantis}, Vélizy-Villacoublay, France
    \\\{1, 5, 6\}@stellantis.com}
    \IEEEauthorblockA{\IEEEauthorrefmark{2}\textit{IMT Atlantique}, Lab-STICC, UMR CNRS 6285, F-29238 Brest, France
    \\\{1, 2, 3, 4\}@imt-atlantique.fr}
}

\maketitle

\begin{abstract}
Structured pruning is a popular method to reduce the cost of convolutional neural networks, that are the state of the art in many computer vision tasks. However, depending on the architecture, pruning introduces dimensional discrepancies which prevent the actual reduction of pruned networks. To tackle this problem, we propose a method that is able to take any structured pruning mask and generate a network that does not encounter any of these problems and can be leveraged efficiently. We provide an accurate description of our solution and show results of gains, in energy consumption and inference time on embedded hardware, of pruned convolutional neural networks. 
\end{abstract}

\begin{IEEEkeywords}
Deep Learning, Compression, Pruning, Energy, Inference, GPU
\end{IEEEkeywords}

\section{Introduction}

Deep neural networks are at the state of the art in many domains, such as computer vision. For instance, convolutional neural networks are used to tackle different tasks such as classification~\cite{ILSVRC15} or semantic segmentation~\cite{ronneberger2015u}. However, their cost in energy, memory and latency is prohibitive on embedded hardware, and this is why many works focus on reducing their cost to fit targets with limited resources~\cite{chen2016deep}.

The field of deep neural networks compression counts multiple types of method, such as quantization~\cite{courbariaux2015binaryconnect} or distillation~\cite{hinton2015distilling}. The one we focus on in this article is pruning~\cite{han2015learning}, that involves removing unnecessary weights from a network.  Pruning is a popular technique that presents many challenges, including that of finding the most adequate type of sparsity to be leveraged on hardware~\cite{ma2021non}.

To focus on the theoretical approach of studying the impact of removing weights from the network's function on its accuracy, many papers only remove weights by putting their value to zero. However, this does not reduce the cost of networks and only provides a rough estimate of network compression in terms of memory. Leveraging pruning to get gains on hardware is actually not a trivial task. Pruning isolated weights~\cite{han2015learning} (``non-structured pruning'') produces sparse matrices, that are difficult to accelerate~\cite{ma2021non}. Pruning entire convolution filters\textbf{} (a.k.a. ``structured pruning'') is more easily exploitable, but the input and output dimensions of layers are altered, which can induce many problems in networks, especially those including long-range dependencies between layers~\cite{he2016deep}. The solution to this problem is, almost always, either not mentioned, or circumvented by constraining pruning into targeting only layers that do not induce problems~\cite{li2016pruning}. However, these constraints are expected to reduce the efficiency of pruning.

In this paper we propose a solution to reduce effectively the size of networks using structured pruning, that were applied a mask using structured pruning. Our method is generic, automatic and reliably produces an effectively pruned network. We demonstrate its ability to operate on networks of any complexity by applying it on both a standard classification network~\cite{he2016deep} on the ImageNet ILSVRC2012 dataset~\cite{ILSVRC15} and on a more complex semantic segmentation network~\cite{sun2019high} trained on CityScapes~\cite{cordts2016cityscapes}. We show that our solution allows gains in energy consumption and inference time on embedded hardware such as the NVIDIA Jetson AGX Xavier embedded~GPU, providing an actual estimate of how structured pruning can be leveraged to reduce energy and latency footprints on a real hardware target.

\section{Related Works}\label{sec:relwo}

\begin{figure*}[ht]
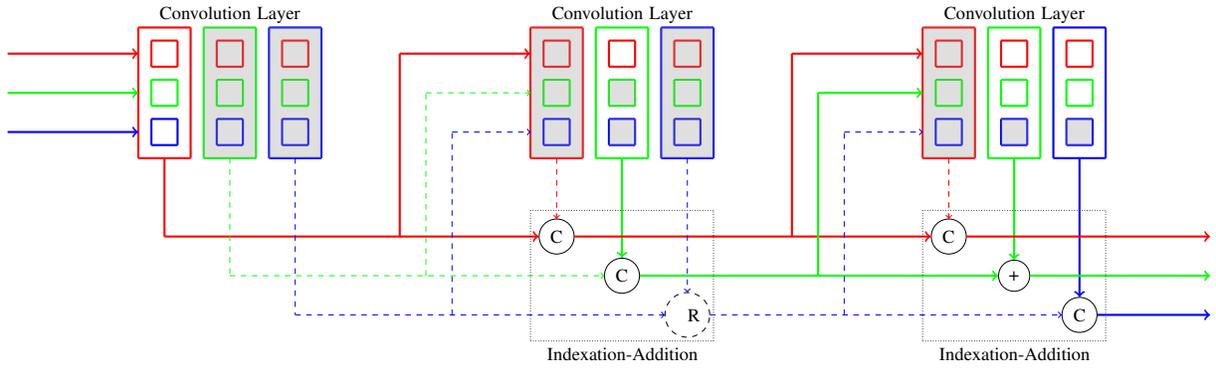

    \centering
    \tikzstyle{none}=[fill=none, draw=none]
    \tikzstyle{circle}=[fill=none, draw=black, shape=circle]
    \tikzstyle{dashed_circle}=[dashed, fill=none, draw=black, shape=circle, text opacity=0]
    
    \tikzstyle{red_line}=[-, draw=red, line width=0.4mm]
    \tikzstyle{green_line}=[-, draw=green, line width=0.4mm]
    \tikzstyle{blue_line}=[-, draw=blue, line width=0.4mm]
    \tikzstyle{red_arrow}=[draw=red, ->, line width=0.4mm]
    \tikzstyle{green_arrow}=[draw=green, ->, line width=0.4mm]
    \tikzstyle{blue_arrow}=[draw=blue, ->, line width=0.4mm]
    \tikzstyle{red_dashed}=[-, draw=red, dashed]
    \tikzstyle{green_dashed}=[-, draw=green, dashed]
    \tikzstyle{blue_dashed}=[-, draw=blue, dashed]
    \tikzstyle{red_dashed_arrow}=[draw=red, ->, dashed]
    \tikzstyle{green_dashed_arrow}=[draw=green, ->, dashed]
    \tikzstyle{blue_dashed_arrow}=[->, draw=blue, dashed]
    \tikzstyle{pruned}=[-, fill={rgb,255: red,150; green,150; blue,150}, opacity=0.3, draw=none]
    \tikzstyle{black_dashed}=[-, densely dotted]
    
    \pgfdeclarelayer{nodelayer}
    \pgfdeclarelayer{edgelayer}
    \pgfsetlayers{nodelayer, edgelayer}
    
    \resizebox{0.9\textwidth}{!}{

    }
    \caption{Illustration of the difficulties when pruning filters in convolutional neural networks. Convolution layers are made of filters, each one outputing a channel (or ``feature map''). Greyed out elements symbolise pruned filters and the kernels to remove to fit the dimensions of inputs (Problem 1). At the end of every residual block, the output of the last layer is summed with the input of the block. If the two tensors are pruned differently (Problem 3), what was an addition is now a mixture of additions (+), concatenations (C) or bypasses (dashed circles) that we call the \textit{indexation-addition} operator (Section~\ref{sec:indexation}). The consequence is that the final number of channels cannot be predicted solely from a particular layer in the network, but must be deduced by taking into account all the dependencies (Problem 2).}
    \label{fig:problems}
\end{figure*}

Originally designed to improve generalization of neural networks~\cite{lecun1989optimal}, pruning is now a popular method to reduce their memory or computational footprints. The most basic form of pruning involves masking out weights of least magnitude in a non-structured way~\cite{han2015learning}. This method does not reduce the size of the parameters' tensors, but instead the introduced zeroes help compressing the network weights through encoded schemes~\cite{han2015deep}. However, getting any type of speed-up out of this method is difficult on most hardware~\cite{ma2021non}.

To better leverage pruning on hardware, many methods instead apply ``structured pruning'', that usually involves pruning whole neurons, \textit{i.e.} filters in the case of convolution layers~\cite{li2016pruning}. Other types of structured pruning exist, such as ``filter shape pruning''~\cite{wen2016learning} and this is why we will favor the ``filter pruning'' denomination to avoid ambiguity. Weight pruning and filter pruning are the two most popular types of pruning structures.

When pruning any type of structure, two aspects have to be tackled: 1) how to identify elements to prune and 2) how to prune them. The first issue can be solved using various types of pruning criteria. In the case of non-structured pruning, the magnitude of weights~\cite{han2015learning} or their gradient~\cite{molchanov2016pruning} are two popular criteria. When pruning filters, these criteria can be extended to either their norm over a filter~\cite{li2016pruning} or a proxy that accounts for the whole filter's importance, for example the multiplicative learned weight included in batch-normalization layers~\cite{liu2017learning}. These criteria can be applied in two different ways: either they are used to identify the same (or a pre-determined) amount of weights/filters to remove in all layers (local pruning) or the target is set globally and the criterion is applied to all layers at the same time (global pruning).

Concerning the second issue, many popular methods apply a simple framework~\cite{han2015deep}: training the network, pruning a given proportion of weights by masking them away, fine-tuning the network and repeating the last two steps multiple times until a target pruning rate is reached. Other methods can involve a more progressive approach~\cite{he2018soft} that can include a regrowing mechanism~\cite{mocanu2018scalable}. Some techniques propose a more continuous way to prune weights, for example by applying them a penalty during training~\cite{tessier2022rethinking}.

\section{Method}

\begin{figure}[ht]
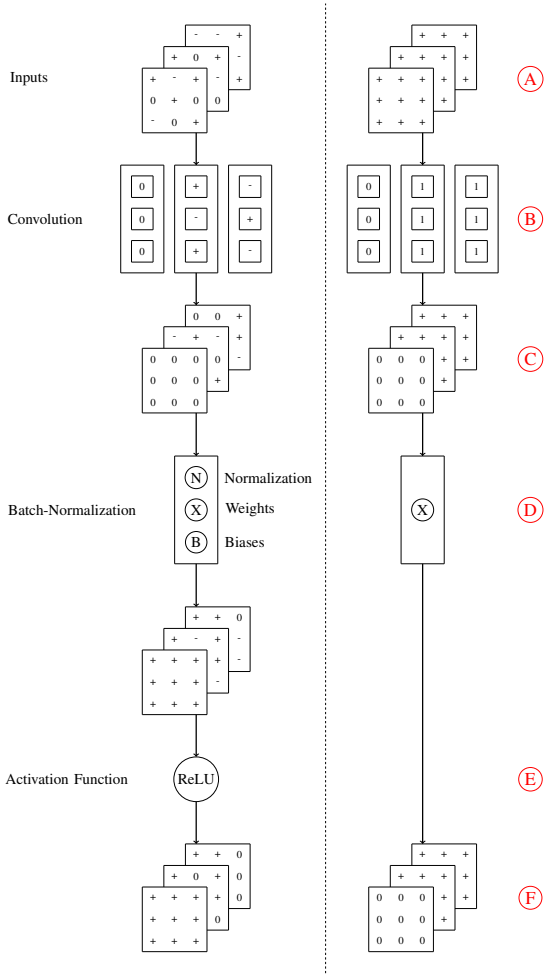

    \centering
    

    \tikzstyle{none}=[fill=none, draw=none, shape=circle]
    \tikzstyle{circle}=[fill=none, draw=black, shape=circle]
    \tikzstyle{red_circle}=[fill=none, draw=red, shape=circle]
    
    \tikzstyle{every node}=[font=\Large]
    
    \tikzstyle{black}=[-]
    \tikzstyle{Arrow}=[->, line width=0.6mm]
    \tikzstyle{dashed_line}=[-, dashed]
    
    \pgfdeclarelayer{nodelayer}
    \pgfdeclarelayer{edgelayer}
    \pgfsetlayers{nodelayer, edgelayer}
    
    \resizebox{0.4\textwidth}{!}{

    }
    \caption{Illustration of the proposed method to identify disconnected weights, with the original network on the left and the modified version on the right. (A) Input tensors are uniform to avoid unwanted null values, (B) weights of layers are replaced with their mask, therefore (C) the output only contain null values if a filter is pruned. (D) Normalization and biases are removed to keep null values null and (E) activation functions are removed not to add extra ones. The final output (F) allows deducing which filters are pruned.}
    \label{fig:modif}
\end{figure}

\subsection{Consequences of Structured Pruning}\label{sec:consequences}

In Section~\ref{sec:relwo}, we explained what is structured pruning. In order to present the problems it can induce, as well as the solutions we propose, we need to introduce some notations.

Let $\mathcal{N}$ be a convolutional neural network. For the sake of convenience, we will consider that it is only made of convolutional layers $l^i$, whose input and output dimensions are $f_{in}^i$ and $f_{out}^i$. Each convolution contains $f_{out}^i\times f_{in}^i\times k^i_h\times k^i_w$ weights $\mathbf{w}^i$ (with $k^i_h\times k^i_w$ the size of the layer's kernel) and $f_{out}^i$ biases $\mathbf{b}^i$. A filter corresponds to the $f_{in}^i\times k^i_h\times k^i_w$ weights and one bias that produce one of the $f_{out}^i$ channels in the output feature maps. Each of these layers operates on feature maps of size $f_{in}^i\times h^i\times w^i$ with $h^i\times w^i$ the resolution of the feature maps. In the case of networks such as ResNet~\cite{he2016deep} or HRNet~\cite{sun2019high}, different layers can take the same feature maps as an input and multiple feature maps can be summed together. This simplified presentation is sufficient to expose the problems induced by global pruning.

\paragraph{Problem 1} Pruning filters reduces the output dimension $f_{out}$ of a layer. Therefore, the dimension of its output is different and the input dimension $f_{in}$ of the following layers must be adapted. This problem is well-known in the literature~\cite{li2016pruning} and easy to solve in simple networks.

\paragraph{Problem 2} Residual connections~\cite{he2016deep} can introduce long-range dependencies and, therefore, identifying all the layers impacted by the change in dimension can be difficult. This problem is usually solved by avoiding pruning layers involved in such dependencies~\cite{li2016pruning}, but this solution is suboptimal.

\paragraph{Problem 3} Residual connections~\cite{he2016deep} usually involve summing together feature maps, that must therefore be of same dimensions, which is not the case anymore after global pruning. In the case of local pruning, feature maps are of the same dimensions, but the same mask may not have been applied on both feature maps, and summing together channels that are meant to be summed together produces a tensor of higher dimensions. This problem is less discussed in the literature and mostly solved using custom operations to manually adapt dimensions of feature maps~\cite{he2017channel}.

These three problems, illustrated in Figure~\ref{fig:problems}, are either eluded or not solved in the literature, even though most papers deal with ResNet-based architectures that are causing all of the three. Some expertise allows manually figuring out dependencies in such networks, but the complexity can get out of hand in the case of networks such as HRNets~\cite{sun2019high}. Indeed, missing any of these problems makes the networks impossible either to run efficiently or to run at all on hardware. This is the reason why we propose a method that can automatically and reliably produce pruned networks that can be ran efficiently on hardware.

\subsection{Generalizing Operators to Handle a Subset of Channels}~\label{sec:scatter}

The first step of our method is to make sure a given network is robust to pruning. Indeed, networks such as ResNets or HRNets contain operations that are applied to outputs of multiple layers. In such cases, the involved tensors must be of the same dimension, which may not be the case anymore after pruning. In the case of ResNets and HRNets, all operations of these types are additions of two tensors, such as those at the end of every residual connection. This means that we can tackle this problem by replacing additions with a generalized operator able to handle missing filters in any of its inputs.

To this mean, we replace additions with a new \textit{indexation-addition} operation, with $\mathbf{a}$ and $\mathbf{b}$ the tensors to sum, that contain respectively $n^a$ and $n^b$ channels, $\mathbf{i}^a$ and $\mathbf{i}^b$ two lists of indices and the output tensor $\mathbf{c}$, that contains $n^c$ channels, defined in Equation~\eqref{eq:eqn}:
\begin{align}
    \forall k \in \llbracket 1; n^c\rrbracket,
    \mathbf{c}_k = &\begin{cases}
      \mathbf{a}_{\mathbf{i}^a_k}, & \text{if}\ \mathbf{i}^a_k \in \llbracket 1; n^a\rrbracket \\
      \varnothing, & \text{otherwise}
    \end{cases}   \label{eq:eqn}\\
    + &\begin{cases}
      \mathbf{b}_{\mathbf{i}^b_k}, & \text{if}\ \mathbf{i}^b_k \in \llbracket 1; n^b\rrbracket \\
      \varnothing, & \text{otherwise}\end{cases}\nonumber
\end{align}

If $n^a = n^b$, $\mathbf{i}^a = \left[ 1, 2, \dots, n^a \right]$ and $\mathbf{i}^b = \left[ 1, 2, \dots, n^b \right]$, this \textit{indexation-addition} operation is purely equivalent to an element-wise addition. Properly parameterized by adequate $\mathbf{i}^a$ and $\mathbf{i}^b$, this operation allows leveraging any type of filter pruning. It is however necessary to find the right $\mathbf{i}^a$ and $\mathbf{i}^b$ and we provide a solution in Section~\ref{sec:indexation}. Figure~\ref{fig:problems} illustrates how our solution relates to the problems mentioned in Section~\ref{sec:consequences} and provides a simple way to view how it can behave like a mix of additions and concatenations.

\subsection{Automatic Adaptation of Networks}~\label{sec:AAN}

Once the network is prepared for pruning by the introduction of this new \textit{indexation-addition} operation to fit any distribution of the sparsity induced by pruning, the next step of the method is to identify automatically all dependencies between filters, kernels, biases or any sort of weights in the network. In summary, it is necessary to search for all the parts of the network that are disconnected when removing filters. 

To identify all parameters whose contribution in a network's function is null, one can use its gradient over, for example, a mini-batch from the training data. Indeed, provided this mini-batch is a satisfying approximation of the network's domain of definition, a null gradient means that the network's function is null relatively to the involved weights, or at least constant in the case of biases. However, for our use-case, this is insufficient: not only does it not allow removing disconnected biases that still produce constant outputs that somehow contribute to the function, but it may also identify some isolated weights as pruned in a non-structured way, while it is not possible to leverage them.

This is why we instead operate on an architectural abstraction of the network, which is a copy of it that received three modifications that are illustrated in Figure~\ref{fig:modif}:
\begin{itemize}
    \item Its biases are removed to prevent them from adding a constant output that makes some disconnected/useless weights downstream have a non-null gradient.
    \item Its activation functions, and other non-linear operations such as normalization, are removed, so that a non-null input of a layer cannot produce a null output and  gradient.
    \item The value of its weights are replaced by the value of the mask, made either of zeros or ones, so that, when fed with an input filled with non-null values of the same sign, the output cannot contain null values if it is not because of null, masked out weights.
\end{itemize}
Because of these modifications, a single input filled with non-null values of the same sign is enough to identify all disconnected weights. Indeed, this network behaves like a purely linear and positive function and any null gradient in its parameters can only be due to a null function that can be removed. Weights, identified as disconnected in this copy network, are then removed from the original network.

\subsection{Automatic Indexation}\label{sec:indexation}

To deduce automatically the right $\mathbf{i}^a$ and $\mathbf{i}^b$ defined in Section~\ref{sec:scatter}, we add another modification to the copy network described in Section ~\ref{sec:AAN}:
we apply an \textit{identity convolution} to the two tensors before summing them together. This \textit{identity convolution} has weights of shape $n\times n \times 1 \times 1$ (with $n$ the number of channels in the input tensor) whose values equates that of an identity matrix. 

The gradient of the weights of this \textit{identity convolution} allows deducing the corresponding list of indices.
Indeed, once the null rows and columns of its weights are removed, the output dimensions are the same for both tensors to be summed while the input dimension matches that of the input tensors after pruning. The zero and non-zero remaining coefficient allows deducing how to map the input and output channels.

\subsection{Summary of the Method}

Here are all the steps to follow to apply our method:
\begin{algorithm}
\caption{Summary of the Method}\label{alg:meth}
\begin{algorithmic}[1]
    \State train the network $\mathcal{N}$
    \State generate the pruning mask $\mathbf{m}$ that masks out filters
    \State create a copy $\mathcal{N}'$ of the network
    \State remove all biases $\mathbf{b}$ from $\mathcal{N}'$
    \State remove all activation functions and normalization from $\mathcal{N}'$
    \State replace the weights $\mathbf{w}$ of $\mathcal{N}'$ by $\mathbf{m}$
    \State insert the \textit{identity convolutions} where needed in $\mathcal{N}'$
    \State generate an input tensor $\mathbf{x}$, of adequate size, filled with ones and run $\mathcal{N}'(\mathbf{x})$
    \State compute $\frac{d\mathcal{N}'}{d\mathbf{w}}(\mathbf{x})$
    \State generate the new pruning mask $\mathbf{m}'$ that masks away all weights whose gradient is null in $\mathcal{N}'$
    \State apply $\mathbf{m}'$ to $\mathcal{N}$ and mask away biases whose weights are pruned
    \State deduce from the mask of the \textit{identity convolutions} the right $\mathbf{i}^a$ and $\mathbf{i}^b$ to replace additions with \textit{indexation-addition} operations where needed
\end{algorithmic}
\end{algorithm}

The method, summed up in Algorithm~\ref{alg:meth}, solves all problems presented in Section~\ref{sec:consequences}. It allows pruning a network and then generating its nearest equivalent whose dimensions are consistent and that can be leveraged on hardware. Since our method not only removes weights of null contribution but also biases whose gradient is constant, the function of the network is not preserved. However, the impact on accuracy is negligible and detailed in our experiments in Section~\ref{sec:accuracy}.

\section{Experiments}

In our experiments, we will first detail the impact of our method on both the accuracy of the network and the evaluation of its compression rate. Then we will demonstrate how the networks, whose type of sparsity usually prevents running inference, can be leveraged efficiently on resource-limited hardware. Our source code is available at: https://github.com/HugoTessier-lab/Neural-Network-Shrinking.git

\subsection{Training conditions}

\paragraph{ImageNet} We trained ResNet-50~\cite{he2016deep} on the ImageNet ILSVRC2012 image classification dataset~\cite{ILSVRC15} for 90 epochs with a batch-size of 170 and a learning rate of 0.01 reduced by 10 every 30 epochs. We used the SGD optimizer with weight decay set to $1\cdot10^{-4}$ and momentum set to 0.9.

\paragraph{Cityscapes} We trained the HRNet-48 network~\cite{sun2019high} on the Cityscapes semantic segmentation dataset~\cite{cordts2016cityscapes} for 200 epochs with a batch size of 10 and a learning rate of 0.01 reduced by $(1-\frac{current\_epoch}{epochs})^2$ at each epoch. We used the RMI loss~\cite{zhao2019region} and the SGD optimizer with weight decay set to $5\cdot10^{-4}$ and momentum set to 0.9. During training, images are randomly cropped and resized, with a scale of $[0.5, 2]$, to $3\times512\times1024$. Data augmentation involves random flips, random Gaussian blur and color jittering.

\paragraph{Pruning} We prune networks following the method of Liu~et~al.~\cite{liu2017learning}: pruning is divided in three iterations, with a linearly growing proportion of removed filters until the final pruning rate is matched. At each iteration, filters are masked out depending on the magnitude of the weight of their batch-normalization layer. After each iteration, ResNet-50 fine-tuned during 10 epochs and HRNet-48 during 20 epochs. The method of Liu~et~al.~\cite{liu2017learning} also implies penalizing weights of batch-normalization layers with a smooth-$\mathcal{L}_1$ norm, with an importance factor of  $\lambda = 10^{-5}$ for ResNet-50 and $\lambda = 10^{-6}$ for HRNet-48.

\subsection{Impact on Accuracy and Compression Rate}\label{sec:accuracy}

\begin{figure}[ht]
    \centering
        	\begin{tikzpicture}
    \begin{scope}[scale=0.9]
    \begin{axis}[
    ylabel=Top-1 accuracy (\%),
    mark size = 1pt,
    xlabel=Pruning Rate (\%),
    ylabel shift = -6 pt,
    width=5cm,
    height=5cm,
    legend entries={Before, After},
    legend to name=named,
    legend columns=-1,
    legend style = {draw=none},
    xtick pos=left,
    ytick pos=left
    ]
    
    \addplot coordinates 
    {
    	(0,75.7)
    	(10,74.642)
    	(20,74.672)
    	(30,73.972)
    	(40,73.188)
    	(50,70.57)
    	(60,66.05)
    	(70,53.022)
    	(80,33.142)
    	(90,1.188)
    };
    \addplot coordinates
    {
    	(0,75.7)
    	(2.82, 74.642)
    	(5.76, 74.672)
    	(12.07, 73.972)
    	(25.53, 73.188)
    	(41.88, 70.57)
    	(61.12, 63.478)
    	(78.16, 53.068)
    	(88.64, 33.156)
    	(92.14, 1.152)
    	
    };

    \end{axis}
    \end{scope}
    \end{tikzpicture}
    \begin{tikzpicture}
    \begin{scope}[scale=0.9]
    \begin{axis}[
    ylabel=mIoU (\%),
    mark size = 1pt,
    xlabel=Pruning Rate (\%),
    ylabel shift = -6 pt,
    width=5cm,
    height=5cm,
    xtick pos=left,
    ytick pos=left
    ]
    
    \addplot coordinates 
    {
    	(0 , 77.0)
    	(10, 77.07639)
    	(20, 76.5922)
    	(30, 74.786) 
    	(40, 71.558)
    	(50, 62.1228)
    	(60, 37.29596)
    	(70, 16.37308)
    	(80, 7.749) 
    	(90, 5.96049)
    };
    \addplot coordinates
    {
    	(0 , 77.0)
    	(6.26, 77.07639)
    	(16.16, 76.5922)
    	(28.94, 74.786) 
    	(43.06, 71.558)
    	(57.26, 62.1228)
    	(70.9, 37.29596)
    	(82.67, 16.37308)
    	(92.15, 7.793) 
    	(97.67, 6.24168)

    };

    \end{axis}
    \end{scope}
    \end{tikzpicture}
    \ref{named}
    
    \caption{For ResNet-50 on ImageNet (left) or HRNet-48 on Cityscapes (right): accuracy depending on pruning rate, either in terms of proportion of pruned filters (blue) or remaining parameters after application of our method~(red).}
    \label{fig:accuracy}
\end{figure}
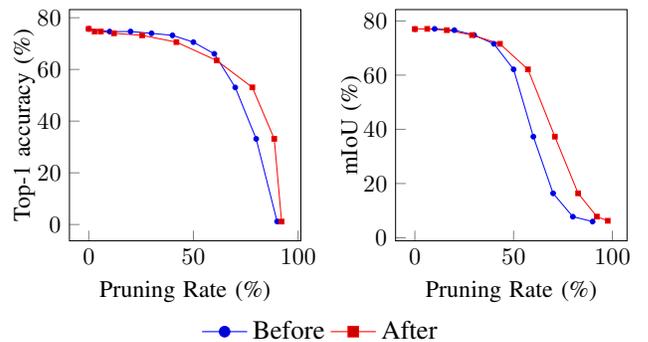

In our experiments, we reported mostly no difference in accuracy before and after applying our method, as it can be seen in Figure~\ref{fig:accuracy}. This implies that the parameters removed by our method, that did not have a null contribution to the function, such as the remaining biases mentioned in Section~\ref{sec:AAN}, might have had a negligible impact on the network's accuracy. The only outliers are points where accuracy is already severely decreased, for example the accuracy of ResNet-50 pruned at 60\% that goes from 66.05\% to 63.478\%, while the baseline is at 75.7\%.

In Figure~\ref{fig:accuracy} we also show the trade-off between accuracy and two types of pruning rate: one defined as the proportion of removed filters, which is a widespread target criterion in the literature, and one defined as the exact count of remaining parameters in the network once our method has been applied. We see that using the percentage of removed filter is not faithful to the actual compression rate of the network. The actual trade-off is more advantageous once our method has been applied to both purge the network from useless weights and get a faithful estimation of all eliminated weights.

\begin{figure}[ht]
    \centering
        	\begin{tikzpicture}
    \begin{scope}[scale=0.9]
    \begin{axis}[
    ylabel=Param. Comp.,
    mark size = 1pt,
    xlabel=Filters Comp.,
    ylabel shift = -6 pt,
    width=5cm,
    height=5cm,
    xtick pos=left,
    ytick pos=left,
    xmode=log,
    ymode=log
    ]
    
    \addplot coordinates 
    {
    	(1.111,1.02901831653)
    	(1.25, 1.06112054329)
    	(1.428, 1.13726828159) 
    	(1.67, 1.34282261313)
    	(2, 1.72057811425)
    	(2.5, 2.57201646091)
    	(3.33,4.57875457875)
    	(5, 8.80281690141) 
    	(10, 12.7226463104)
    };

    \end{axis}
    \end{scope}
    \end{tikzpicture}
    \begin{tikzpicture}
    \begin{scope}[scale=0.9]
    \begin{axis}[
    ylabel=Param. Comp.,
    mark size = 1pt,
    xlabel=Filters Comp.,
    ylabel shift = -8 pt,
    width=5cm,
    height=5cm,
    xtick pos=left,
    ytick pos=left,
    xmode=log,
    ymode=log
    ]
    
    \addplot coordinates 
    {
    	(1.111, 1.067)
    	(1.25, 1.19)
    	(1.428, 1.407) 
    	(1.67, 1.756)
    	(2, 2.339)
    	(2.5, 3.436)
    	(3.33, 5.77)
    	(5, 12.7388) 
    	(10, 42.918)
    };

    \end{axis}
    \end{scope}
    \end{tikzpicture}
    \caption{For ResNet-50 on ImageNet (left) or HRNet-48 on Cityscapes (right): relation between the estimated compression rate in terms of pruned filters (x-axis) and remaining parameters after reducing the network using our method (\mbox{y-axis}).}
    \label{fig:comprate}
\end{figure}
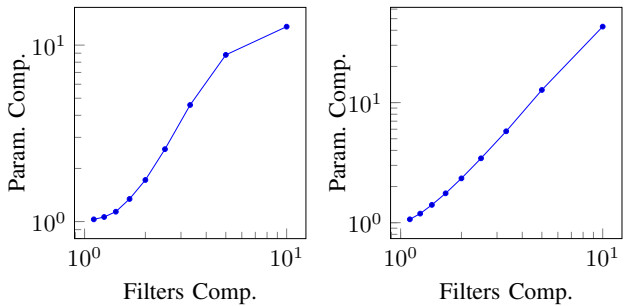

In Figure~\ref{fig:comprate}, we compare the compression rate (\textit{i.e.} $\frac{100\%}{100\% - pruning\_rate\%}$) in terms of removed filters or removed parameters, \textit{i.e.} before and after our method. The relationship between the two measures seem to depend on the involved architecture and we expect it to depend on the pruning criterion~too.

\subsection{Impact on Hardware}

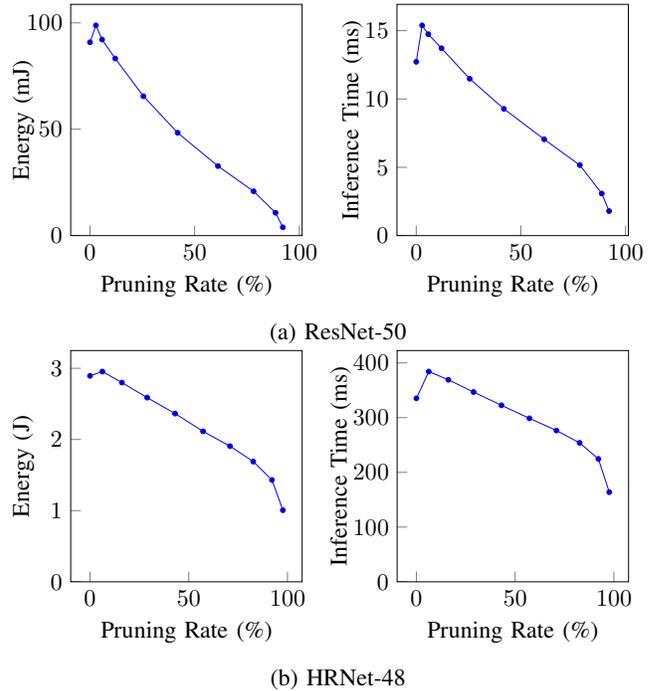
\begin{figure}[ht]
    \centering
    \begin{subfigure}{1\linewidth}
    	
    	\begin{tikzpicture}
    	\begin{scope}[scale=0.9]
    	\begin{axis}[
    	ymin=0,
    	ylabel=Energy (mJ),
    	mark size = 1pt,
    	xlabel=Pruning Rate (\%),
    	ylabel shift = -10 pt,
    	width=5cm,
    	height=5cm,
    	xtick pos=left,
    	ytick pos=left
    	]
    	
    	\addplot coordinates 
    	{
    		(0, 90.8498)
    		(2.82, 98.8062)
    		(5.76,  92.1657)
    		(12.07, 83.2253)
    		(25.53, 65.5189)
    		(41.88, 48.3323)
    		(61.12, 32.6885)
    		(78.16, 20.8484)
    		(88.64, 10.754)
    		(92.14, 3.8987)
    	};

    	\end{axis}
    	\end{scope}
    	\end{tikzpicture}
    	\begin{tikzpicture}
    	\begin{scope}[scale=0.9]
    	\begin{axis}[
    	ymin=0,
    	ylabel=Inference Time (ms),
    	mark size = 1pt,
    	xlabel=Pruning Rate (\%),
    	ylabel shift = -6 pt,
    	width=5cm,
    	height=5cm,
    	xtick pos=left,
    	ytick pos=left
    	]
    	
    	\addplot coordinates
    	{
    		(0, 12.725573539733887)
    		(2.82, 15.387354564666748)
    		(5.76,  14.741002798080444)
    		(12.07, 13.71607756614685)
    		(25.53, 11.483997559547424)
    		(41.88, 9.277763986587524)
    		(61.12, 7.050588488578796)
    		(78.16, 5.159873986244202)
    		(88.64, 3.0823596000671387)
    		(92.14, 1.7969604015350342)
    		
    	};
    	
    	\end{axis}
    	\end{scope}
    	\end{tikzpicture}
    	\caption{ResNet-50}
    \end{subfigure}

    \begin{subfigure}{1\linewidth}
    	\begin{tikzpicture}
    	\begin{scope}[scale=0.9]
    	\begin{axis}[
    	ymin=0,
    	ylabel=Energy (J),
    	mark size = 1pt,
    	xlabel=Pruning Rate (\%),
    	ylabel shift = 0 pt,
    	width=5cm,
    	height=5cm,
    	xtick pos=left,
    	ytick pos=left
    	]
    	
    	\addplot coordinates 
    	{
    		(0 , 2.894034)
    		(6.26, 2.953858)
    		(16.16, 2.799311)
    		(28.94, 2.588446) 
    		(43.06, 2.364164)
    		(57.26, 2.114278)
    		(70.9, 1.906126)
    		(82.67, 1.689524)
    		(92.15, 1.430929)
    		(97.67, 1.005973) 
    	};

    	\end{axis}
    	\end{scope}
    	\end{tikzpicture}
    	\begin{tikzpicture}
    	\begin{scope}[scale=0.9]
    	\begin{axis}[
    	ymin=0,
    	ylabel=Inference Time (ms),
    	mark size = 1pt,
    	xlabel=Pruning Rate (\%),
    	ylabel shift = -6 pt,
    	width=5cm,
    	height=5cm,
    	xtick pos=left,
    	ytick pos=left
    	]
    	
    	\addplot coordinates 
    	{
    		(0 , 335.166707277298)
    		(6.26, 384.0981516838074)
    		(16.16, 369.1493010520935)
    		(28.94, 346.7119140625) 
    		(43.06, 322.4898753166199)
    		(57.26, 298.6627109050751)
    		(70.9, 276.4353265762329)
    		(82.67, 253.81475520133972)
    		(92.15, 224.5712697505951)
    		(97.67, 163.77321767807007) 
    	};
    	
    	\end{axis}
    	\end{scope}
    	\end{tikzpicture}
    	\caption{HRNet-48}
    \end{subfigure}
    
    \caption{Energetic consumption and inference time of ResNet-50 and HRNet-48, depending on the pruning rate in terms of parameters, on NVIDIA Jetson AGX Xavier in the ``30W All'' mode, using JetPack SDK 5.0 and ONNX Runtime 1.12.0 running with the TensorRT execution provider. Results are averaged over 10k inferences with inputs of size $(1\times 3\times 224\times 224)$ after 1k runs of warm-up for ResNet-50 and 1k inferences with inputs of size $(1\times 3\times 512\times 1024)$ after 100 runs of warm-up for HRNet-48.}
    \label{fig:consumption}
\end{figure}


To measure the inference time and energetic consumption of pruned networks on NVIDIA Jetson AGX Xavier in the ``30W All'' mode, we first converted our networks to ONNX, that is a format that can be handled by many frameworks on most hardware. The \textit{indexation-addition} operations were implemented using \textit{ScatterND} and \textit{transpose} operators. \textit{ScatterND} allows operating on slices in tensors and transpositions allow operating specifically on channels, while \textit{Scatter} is element-wise and requires storing a cumbersome array of indices. Before summation, both tensors need to be scattered into a temporary tensor, that is instantiated dynamically. 
We used the JetPack SDK 5.0, with CUDA 11.4.14, cuDNN 8.3.2, TensorRT 8.4.0 EA and ONNX Runtime 1.12.0. Energetic consumption was given using the tegrastats utility. Inference on ResNet-50 is run with an input of size $(1\times 3\times 224\times 224)$ and HRNet-48 with one of size $(1\times 3\times 512\times 1024)$. ONNX Runtime was used with the TensorRT execution provider, that turned out to be the one that gave the best inference time.

Figure~\ref{fig:consumption} provides results for ResNet-50 on ImageNet and HRNet-48 on Cityscapes. Both show similar tendencies: at first, the extra cost of \textit{indexation-addition} operations takes a toll on the efficiency of pruning, but after that initial jump, the cost of networks, either in terms of energy consumption or inference time, decreases significantly. This shows that, although a better implementation of the \textit{indexation-addition} operations would be beneficial, our current solution is enough for free and unconstrained structured pruning to be cost effective.
Therefore, we can say that it is possible to leverage efficiently any type of filter pruning in even complex deep convolutional neural networks.

\section{Discussion}

Three observations can be drawn from our experiments: 1) our method allows a more reliable measurement of the count of remaining parameters in the network, as can be seen in Figure~\ref{fig:accuracy}, 2) the relation between this accurate pruning rate and inference time or energy consumption is non-linear and 3) the cost introduced by our custom operators is not negligible and makes the least pruned networks cost more than non-pruned ones, as can be seen in Figure~\ref{fig:consumption}.

The first two observations show that our method is a useful tool to better study the efficiency of unconstrained filter pruning. Indeed, it produces a network in which the vast majority of remaining parameters are guaranteed to contribute to the function, with the marginal exception of some isolated weights that may be inactive by accident. Therefore, it is now possible to directly measure the accuracy-to-energy or accuracy-to-latency trade-off, which provide a more relevant insight into the impact of pruning on hardware than a more theoretical accuracy-to-parameters trade-off. Since this is not the focus of this article, we did not provide such an analysis and did not choose the pruning method that gave the absolute best possible performance. This will be the focus of future contributions.
This ability to provide a more faithful compression rate than the naive rate of removed filters also allows better controlling the growth of pruning rate between pruning iterations. This is likely to help improving performance and avoiding to remove entire layers by accident, which is called \textit{layer collapse}~\cite{tanaka2020pruning}.

Concerning the last observation, finding the best implementation of the custom operators, necessary to run pruned networks, obviously requires further investigation. Using trtexec, we did the profiling of the operators of the HRNet-48, with 10\% of the filters pruned and 6.26\% of removed parameters, which is the HRNet-48 with the highest inference time. It turned out that the ``Foreign Nodes'' generated by TensorRT, that contain the \textit{ScatterND} we used for our \textit{indexation-addition} operations, are responsible for 14.8\% of the total inference time. When substracting the cost of these nodes from the network's total average time of 369.8ms according to trtexec, the remaining inference time is of 314.3ms, which is actually lower than that of the non-pruned network, which is of 318.7ms. This means that if an optimized implementation of the operators allowed their cost to be negligible, it would make pruning a lot more beneficial, even at low pruning rates.

\section{Conclusion}

We have proposed an efficient and generic way to leverage any type of filter pruning in deep convolutional neural networks. Indeed, even though removing filters in a network can trigger a certain array of problems that can even prevent running its inference, our solution is able to tackle them and generates functional pruned networks that can be run efficiently on hardware. Our experiments, even though they show that our current ONNX implementation has a non-negligible cost, demonstrate that unconstrained filter pruning can be cost-effective.

\bibliographystyle{plain}
\bibliography{sips_paper}

\begin{thebibliography}{10}

\bibitem{chen2016deep}
Chun-Fu Chen, Gwo~Giun Lee, Vincent Sritapan, and Ching-Yung Lin.
\newblock Deep convolutional neural network on ios mobile devices.
\newblock In {\em 2016 IEEE International Workshop on Signal Processing Systems
  (SiPS)}, pages 130--135. IEEE, 2016.

\bibitem{cordts2016cityscapes}
Marius Cordts, Mohamed Omran, Sebastian Ramos, Timo Rehfeld, Markus Enzweiler,
  Rodrigo Benenson, Uwe Franke, Stefan Roth, and Bernt Schiele.
\newblock The cityscapes dataset for semantic urban scene understanding.
\newblock In {\em Proceedings of the IEEE conference on computer vision and
  pattern recognition}, pages 3213--3223, 2016.

\bibitem{courbariaux2015binaryconnect}
Matthieu Courbariaux, Yoshua Bengio, and Jean-Pierre David.
\newblock Binaryconnect: Training deep neural networks with binary weights
  during propagations.
\newblock {\em Advances in neural information processing systems}, 28, 2015.

\bibitem{han2015deep}
Song Han, Huizi Mao, and William~J Dally.
\newblock Deep compression: Compressing deep neural networks with pruning,
  trained quantization and huffman coding.
\newblock {\em arXiv preprint arXiv:1510.00149}, 2015.

\bibitem{han2015learning}
Song Han, Jeff Pool, John Tran, and William Dally.
\newblock Learning both weights and connections for efficient neural network.
\newblock {\em Advances in neural information processing systems}, 28, 2015.

\bibitem{he2016deep}
Kaiming He, Xiangyu Zhang, Shaoqing Ren, and Jian Sun.
\newblock Deep residual learning for image recognition.
\newblock In {\em Proceedings of the IEEE conference on computer vision and
  pattern recognition}, pages 770--778, 2016.

\bibitem{he2018soft}
Yang He, Guoliang Kang, Xuanyi Dong, Yanwei Fu, and Yi~Yang.
\newblock Soft filter pruning for accelerating deep convolutional neural
  networks.
\newblock {\em arXiv preprint arXiv:1808.06866}, 2018.

\bibitem{he2017channel}
Yihui He, Xiangyu Zhang, and Jian Sun.
\newblock Channel pruning for accelerating very deep neural networks.
\newblock In {\em Proceedings of the IEEE international conference on computer
  vision}, pages 1389--1397, 2017.

\bibitem{hinton2015distilling}
Geoffrey Hinton, Oriol Vinyals, Jeff Dean, et~al.
\newblock Distilling the knowledge in a neural network.
\newblock {\em arXiv preprint arXiv:1503.02531}, 2(7), 2015.

\bibitem{lecun1989optimal}
Yann LeCun, John Denker, and Sara Solla.
\newblock Optimal brain damage.
\newblock {\em Advances in neural information processing systems}, 2, 1989.

\bibitem{li2016pruning}
Hao Li, Asim Kadav, Igor Durdanovic, Hanan Samet, and Hans~Peter Graf.
\newblock Pruning filters for efficient convnets.
\newblock {\em arXiv preprint arXiv:1608.08710}, 2016.

\bibitem{liu2017learning}
Zhuang Liu, Jianguo Li, Zhiqiang Shen, Gao Huang, Shoumeng Yan, and Changshui
  Zhang.
\newblock Learning efficient convolutional networks through network slimming.
\newblock In {\em Proceedings of the IEEE international conference on computer
  vision}, pages 2736--2744, 2017.

\bibitem{ma2021non}
Xiaolong Ma, Sheng Lin, Shaokai Ye, Zhezhi He, Linfeng Zhang, Geng Yuan,
  Sia~Huat Tan, Zhengang Li, Deliang Fan, Xuehai Qian, et~al.
\newblock Non-structured dnn weight pruning--is it beneficial in any platform?
\newblock {\em IEEE Transactions on Neural Networks and Learning Systems},
  2021.

\bibitem{mocanu2018scalable}
Decebal~Constantin Mocanu, Elena Mocanu, Peter Stone, Phuong~H Nguyen,
  Madeleine Gibescu, and Antonio Liotta.
\newblock Scalable training of artificial neural networks with adaptive sparse
  connectivity inspired by network science.
\newblock {\em Nature communications}, 9(1):1--12, 2018.

\bibitem{molchanov2016pruning}
Pavlo Molchanov, Stephen Tyree, Tero Karras, Timo Aila, and Jan Kautz.
\newblock Pruning convolutional neural networks for resource efficient
  inference.
\newblock {\em arXiv preprint arXiv:1611.06440}, 2016.

\bibitem{ronneberger2015u}
Olaf Ronneberger, Philipp Fischer, and Thomas Brox.
\newblock U-net: Convolutional networks for biomedical image segmentation.
\newblock In {\em International Conference on Medical image computing and
  computer-assisted intervention}, pages 234--241. Springer, 2015.

\bibitem{ILSVRC15}
Olga Russakovsky, Jia Deng, Hao Su, Jonathan Krause, Sanjeev Satheesh, Sean Ma,
  Zhiheng Huang, Andrej Karpathy, Aditya Khosla, Michael Bernstein,
  Alexander~C. Berg, and Li~Fei-Fei.
\newblock {ImageNet Large Scale Visual Recognition Challenge}.
\newblock {\em International Journal of Computer Vision (IJCV)},
  115(3):211--252, 2015.

\bibitem{sun2019high}
Ke~Sun, Yang Zhao, Borui Jiang, Tianheng Cheng, Bin Xiao, Dong Liu, Yadong Mu,
  Xinggang Wang, Wenyu Liu, and Jingdong Wang.
\newblock High-resolution representations for labeling pixels and regions.
\newblock {\em arXiv preprint arXiv:1904.04514}, 2019.

\bibitem{tanaka2020pruning}
Hidenori Tanaka, Daniel Kunin, Daniel~L Yamins, and Surya Ganguli.
\newblock Pruning neural networks without any data by iteratively conserving
  synaptic flow.
\newblock {\em Advances in Neural Information Processing Systems},
  33:6377--6389, 2020.

\bibitem{tessier2022rethinking}
Hugo Tessier, Vincent Gripon, Mathieu L{\'e}onardon, Matthieu Arzel, Thomas
  Hannagan, and David Bertrand.
\newblock Rethinking weight decay for efficient neural network pruning.
\newblock {\em Journal of Imaging}, 8(3):64, 2022.

\bibitem{wen2016learning}
Wei Wen, Chunpeng Wu, Yandan Wang, Yiran Chen, and Hai Li.
\newblock Learning structured sparsity in deep neural networks.
\newblock {\em Advances in neural information processing systems}, 29, 2016.

\bibitem{zhao2019region}
Shuai Zhao, Yang Wang, Zheng Yang, and Deng Cai.
\newblock Region mutual information loss for semantic segmentation.
\newblock {\em Advances in Neural Information Processing Systems}, 32, 2019.

\end{thebibliography}

\end{document}